\RequirePackage[hyphens]{url}
\documentclass[
]{ceurart}

\usepackage{tablefootnote}
\usepackage{footnote}
\makesavenoteenv{tabular}
\usepackage{listings}
\usepackage{todonotes}
\usepackage{hyperref}
\begin{document}

\copyrightyear{2023}
\copyrightclause{Copyright for this paper by its authors.
  Use permitted under Creative Commons License Attribution 4.0
  International (CC BY 4.0).}

\conference{TEXT2KG: International Workshop on Knowledge Graph Generation from Text,
  May 28--June 01, 2023, Hersonissos, Greece}

\title{Exploring In-Context Learning Capabilities of Foundation Models for Generating Knowledge Graphs from Text}


 \author[1]{Hanieh Khorashadizadeh}[%
email=khorashadizadeh@ifis.uni-luebeck.de,
 ]
 \cormark[1]
\fnmark[1]
\author[2]{Nandana Mihindukulasooriya}[%
orcid=0000-0003-1707-4842,
email=nandana@ibm.com,
 ]
\author[3]{Sanju Tiwari}[%
email=tiwarisanju18@ieee.org
 ]
 \author[1]{Jinghua Groppe}[%
 ]
 \author[1]{Sven Groppe}[%
 ]
\address[1]{University of Lübeck, Germany}
\address[2]{IBM Research, Ireland}
\address[3] {Universidad Autonoma de Tamaulipas, Mexico}



\begin{abstract}
Knowledge graphs can represent information about the real-world using entities and their relations in a structured and semantically rich manner and they enable a variety of downstream applications such as question-answering, recommendation systems, semantic search, and advanced analytics. However, at the moment, building a knowledge graph involves a lot of manual effort and thus hinders their application in some situations and the automation of this process might benefit especially for small organizations. Automatically generating structured knowledge graphs from a large volume of natural language is still a challenging task and the research on sub-tasks such as named entity extraction, relation extraction, entity and relation linking, and knowledge graph construction aims to improve the state of the art of automatic construction and completion of knowledge graphs from text.  

The recent advancement of foundation models with billions of parameters trained in a self-supervised manner with large volumes of training data that can be adapted to a variety of downstream tasks has helped to demonstrate high performance on a large range of Natural Language Processing (NLP) tasks. In this context, one emerging paradigm is in-context learning where a language model is used as it is with a prompt that provides instructions and some examples to perform a task without changing the parameters of the model using traditional approaches such as fine-tuning. This way, no computing resources are needed for re-training/fine-tuning the models and the engineering effort is minimal. Thus, it would be beneficial to utilize such capabilities for generating knowledge graphs from text.

In this paper, grounded by several research questions, we explore the capabilities of foundation models such as ChatGPT to generate knowledge graphs from the knowledge it captured during pre-training as well as the new text provided to it in the prompt. The paper provides a qualitative analysis of a set of example outputs generated by a foundation model with the aim of knowledge graph construction and completion. The results demonstrate promising capabilities. Furthermore, we discuss the challenges and next steps for this research work. 

\end{abstract}


\maketitle

\section{Introduction} 

Knowledge Graphs~\cite{hogan2021knowledge} enable us to represent knowledge about a given domain in a semantically rich manner by representing real-world entities as nodes and relations between them as edges. Such knowledge graphs can be built using standards such as RDF(S) and OWL with well-defined semantics allowing systems to perform reasoning to infer more information or query them using structured query languages such as SPARQL.

There are several approaches for constructing Knowledge Graphs based on the source of the knowledge. They can be constructed from structured data, for example, by converting a relational database into a knowledge graph using mappings such as RDB2RDF~\cite{sahoo2009survey}, or using semi-structured data, for example, DBpedia from Wikipedia Infoboxes, or manually, for example, Wikidata using crowdsourcing. Another approach is to construct knowledge graphs from unstructured sources such as a corpus of text using Natural Language Processing (NLP) techniques such as Named Entity Recognition (NER), Relation Extraction, Open Information Extraction, Entity Linking, and Relation Linking. There is a growing interest in the Semantic Web community to explore such approaches as seen from the workshops such as Knowledge Graph Generation From Text (Text2KG)~\cite{2022text2kg}.

In NLP research, the transformer~\cite{vaswani2017attention} neural network architecture has led to significant improvements in many tasks. Language models such as GPT-2/3 \cite{radford2019language, brown2020language}, BERT \cite{kenton2019bert}, Transformer-XL \cite{dai2019transformer}, ELMo \cite{peters2019knowledge}, RoBERTa \cite{liu2019roberta}, ALBERT \cite{lan2019albert} and XLNet\cite{yang2019xlnet} are becoming popular to improve the results and reducing the human intervention in a wide range of tasks such as searching, question answering and sentence classification.

More recently, there is a focus on building foundation models~\cite{bommasani2021opportunities}. The term foundation model is used to describe a model that is trained on a very large corpus of unlabelled data following the self-supervision paradigm and which can be used or adapted to a wide range of downstream tasks. Foundation models generally show very high transfer learning capabilities. Transfer learning~\cite{thrun2012learning} allows models to acquire knowledge from one task (generally with unlabelled data in a self-learning manner) and apply it to another separate task. For example, a model is pre-trained to predict the next word of a sentence but then fine-tuned to perform text summarization or question answering. 

Reinforcement Learning from Human Feedback (RLHF) is also used to further improve these foundation models at scale~\cite{christiano2017deep, stiennon2020learning}. The idea is to fine-tune language models in a way that they can follow a broad range of written instructions. The human-in-the-loop approach is used to provide feedback based on human preferences as a reward to the reinforcement learning setup. Approaches such as InstructGPT~\cite{ouyang2022training} and ChatGPT have used reinforcement learning from human feedback to improve their models significantly. 

There are several approaches for adapting a foundation model, for example, for a task such as the generation of knowledge graphs from text. One is to perform fine tuning~\cite{fine-tuning-paper} (also known as model tuning) for the task of knowledge-graph generation or its sub-tasks such as relation extraction. This will be essentially updating all model parameters with task-specific training. For models in the scale of foundation models, this requires weeks of GPU time. Prompt tuning~\cite{lester-etal-2021-power} or Prefix-Tuning~\cite{Li2021PrefixTuningOC}, an approach that takes relatively less computational power, is to keep the model parameters frozen and only adds some tunable tokens per downstream task to be prefixed to the input text. Finally, prompt design, where the model is used as it is, but the prompt or the input to the model is designed to provide a few examples of the task~\cite{brown2020language}.  

In-context learning~\cite{min-etal-2022-rethinking, xie2021explanation} is about teaching a model to perform a new task only by providing a few demonstrations of input-output pairs at inference time. The model is supposed to understand the task and the type of output required by the context and instructions provided in the input. This requires the least computational power because no training or tuning is involved.

With these recent advancements of large prompting-based language models such as OpenAI's ChatGPT (175B params), Meta's Galactica (120B params), and Google's Bard (137B params) have been released. As these large foundation models are expensive to fine-tune and train, our goal is to check their capabilities for the task of generating knowledge graphs with in-context learning in a prompt.

Nevertheless, we need to perform further studies to understand not only the capabilities but also the limitations of foundation models. For instance, these models hallucinate generating non-factual and nonsensical text in a fluent and confident manner~\cite{ji2022survey}. Furthermore, there are arguments that they are simply memorizing information without properly understanding the meaning and reasoning in a logical manner. Similarly, there is skepticism about the potential biases these models might have based on their training data and their impact on AI ethics and fairness. In extreme cases, these models might even demonstrate harmful behaviors.

These limitations of foundation models motivate using knowledge graphs instead of just only foundation models from now on because knowledge graphs generally contain manually checked facts and knowledge. In addition, knowledge graphs represent the facts in a symbolic manner that they can be inspected and validated by humans. Such knowledge will also enable explainable AI as it can be used to provide plausible explanations to AI model behaviors. Indeed in future work, one might investigate the possibilities of combining the technologies of foundation models and knowledge graphs to complement each other and overcome the limitations of both.

Foundation models inherently contain knowledge acquired from large corpora of text that seems to be only partly available in structured data sources. For instance, there is a vast amount of information available related to the COVID-19 pandemic in unstructured data sources such as research papers, news articles, Wikipedia, etc. but only a small portion of that information is available in knowledge graphs such as Wikidata. 

It seems to be interesting to explore the possibilities of knowledge graph creation and completion based on the foundation models in order to save efforts and hence costs of the traditional ways using natural language processing on large-scale text collections for information not available in structured data.

Given this background, we will perform an initial exploration of the following research questions:
\begin{itemize}
    \item R1: Can we use the knowledge acquired during pre-training of LLMs for Knowledge Graph completion to fill in missing information with a defined ontology with a few examples?
    \item R2: Can we use LLMs to extract facts to generate knowledge graphs from unseen text that is provided  during inference time?
    \item R3: Given an ontology, can we automatically generate prompts for extracting the relevant triples for the purpose of Knowledge Graph construction?
    \item R4: Given a knowledge graph, can we identify the missing information and create prompts to perform Knowledge Graph completion using foundation models
    \item R5: Given a Knowledge Graph  with some false facts, can we use LLMs to check the given Knowledge Graph and determine which facts are not true for the purpose of fact-checking?
    \item R6: What are the capabilities and limitations of models such as ChatGPT for the above scenarios?
    \item R7: What are the disadvantages and risks associated with such an approach? 
\end{itemize}


The rest of the paper is organized as follows: Section 2 provides an overview of the background of the main concepts discussed in this paper. Section 3 illustrates an architecture to position this work in a high-level big picture to motivate where this work would fit it. Section 4 provides a qualitative analysis of the information extraction capabilities of foundation models in the context of knowledge graph generation from text. Section 5 discusses some of the advantages of the proposed approach and its challenges and Section 6 provides some conclusions.

\section{Background}
\subsection{Foundation Models}
Foundation Models are a general class of models for building artificial intelligence (AI) systems and are trained on enormous data by using self-supervision. These models include GPT-3 \cite{brown2020language}, BERT \cite{devlin2018bert} and CLIP \cite{radford2019language}. Foundation models are not new, treated as a general paradigm of AI, and based on self-supervised learning and deep neural networks \cite{bommasani2021opportunities}. Transfer learning \cite{thrun2012learning} and scale are the key elements of foundation models and pre-training is an effective approach of transfer learning. Transfer learning is the base for constructing the foundation model and scale helps to strengthen these models. There are different stages involved in foundation models: data creation, data curation, training, adaptation, and deployment. Data creation is generally a human-centric approach and it is created by humans and most of the created data is about people. After the data creation, it is required to curate into datasets for the training of foundation model on curated datasets. Adaptation is the fourth stage to create a new model based on the foundation model to perform some tasks such as document summarization. Finally, the foundation model needs to be deployed such that it can be used as an AI system.

\subsection{Knowledge Graphs}

Web technologies have revolutionized the way information was delivered and accessed. They have gone through the era of 1.0 and 2.0 and are stepping into the era of 3.0. Web 1.0 provided techniques for quick information publishing and access, and it is a huge collection of content provided by website owners. Web 2.0 brought the interactive capability into Web 1.0 and enabled users to be contributors as well as consumers of web content. In this area, the web is a collection of content, which contains the collective knowledge of the public. However, Web 2.0 is lack of the capability to extract knowledge from its content and the desire to use the collective knowledge hidden in web content gave birth to Semantic Web technology, which is considered an important ability of Web 3.0. Another enabler of Web 3.0 is the vision of Metaverse~\cite{brown2021metaverse} and this aspect will not be discussed here since it is out of the scope of this work.

At most times, the semantic web is the underlying technology of knowledge graphs. Although the knowledge graph \cite{tiwari2021recent} is defined differently by different works, most definitions follow the RDF\footnote{\url{https://www.w3.org/RDF/}} data model of the semantic web. RDF describes knowledge as a collection of triples of subject, predicate and object, where the predicate indicates the relationship between the subject and the object. For example, the piece of information "COVID-19 is a pandemic that broke out in 2019.” can be described as two RDF triples: <covid19, is, pandemic>, <covid19, breakoutInYear, 2019>. A collection of triples can be visualized as a directed graph, where the subject and object are nodes of the graph and the predicate is an edge directed from the subject node to the object node. In a summary, RDF graphs can be seen as knowledge graphs and any knowledge graphs can be transformed into collections of triples. Instead of RDF triples, the knowledge graph has become the widely used term to describe the relationship of entities. On one side, the introduction of Google’s knowledge graph in 2012 contributed to the popularity of the term~\cite{ehrlinger2016towards}, on the other side, knowledge graph does highlight the nature of the data and thus is a more appropriate and also an impressive term. 

Apart from RDF, the semantic web also provides other standards, including SPARQL\footnote{\url{https://www.w3.org/2001/sw/wiki/SPARQL}}, a standard query language for RDF graphs, and  ontology languages such as RDFS\footnote{\url{https://www.w3.org/2001/sw/wiki/RDFS}}, OWL\footnote{\url{https://www.w3.org/OWL/}} and SHACL\footnote{\url{https://www.w3.org/2001/sw/wiki/SHACL}} for defining the structure and vocabulary of RDF data.

\subsection{Knowledge Graph Construction from Natural Languages}

The Semantic Web is ready to define, represent and query knowledge graphs. What is missing is a way to build knowledge graphs from web content. The construction approaches of knowledge graphs have evolved from semantic publishing to machine learning. The change from RDF triples to knowledge graphs is basically just a change of terminology, the evolution of construction approaches is a giant leap because it means a change from a completely handcrafted approach to an automatic one. 


\paragraph{Natural Language Processing (NLP):}
The goal of NLP is to enable machines to understand the meaning of texts like humans. With this capability, machines can do analysis and processing tasks of texts for human beings. NLP has been widely used for machine translation, text classification, and sentiment analysis. An application on the rise is knowledge extraction, which enables the automatic construction of knowledge graphs from a huge collection of web content. NLP has evolved from rule-based to one powered by the learning ability of AI.

Rule-based: Early NLP relied on complex sets of hand-written rules, which are the formal representations of sophisticated linguistic knowledge and common-sense reasoning. Rule-based NLP creates highly precise solutions, but manually defining complicated rules is a difficult and time-consuming task. A large number of rules is needed to perform an NLP task and such rule-based NLP also suffers from the issue of scalability.

Learning-based: Today’s NLP integrates machine learning technology and has the capability to automatically learn complex rules. The learning-based NLP is reaching a new milestone in the automatic understanding and analysis of texts. It uses machine learning algorithms to train language models on large amounts of data in order to get a solution to the given problem. The technology of machine learning enables a model to steadily optimize itself, so the solution will become increasingly accurate. A number of language models have been trained and can be used for downstream NLP tasks, such as BERT, GPT2, ChatGPT3, RoBERTa, ALBERT, ELECTRA, DeBERTa, XLNet, and T5.  While AI academic community is usually of the opinion that state-of-the-art results obtained by just using more data and more computational resources are not research novelties~\cite{rogers2019transformers}, the huge language models like ChatGPT\footnote{\url{https://openai.com/blog/chatgpt}} (with its 175 billion parameters, 300 billion words, 570GB web content, 10K GPUs and the cost of \$4.6 million for a single training session) does demonstrate the potential of AI-powered NLP.


\subsection{Existing Tools for Language Models}
Language models are AI-based models to create and analyze text. These models are trained to predict the next word in the text, speech recognition, spelling correction, etc. Language models are the basis for natural language processing (NLP) activities such as sentiment analysis and speech-to-text. They are generally categorized into two categories: Statistical Language Models and Neural Language Models\footnote{\url{https://medium.com/unpackai/language-models-in-ai-70a318f43041}}. There are several models\footnote{\url{https://www.marktechpost.com/2023/02/22/top-large-language-models-llms-in-2023-from-openai-google-ai-deepmind-anthropic-baidu-huawei-meta-ai-ai21-labs-lg-ai-research-and-nvidia/?amp}} introduced by different groups such as OpenAI, Google, Deepmind, Anthropic, Baidu, Huawei, Meta, AI21 Labs, LG AI Research and NVIDIA as discussed in Table \ref{table:1}.

\begin{table}
\caption{Existing Language Models}
\centering
\label{table:1}
\begin{tabular}{lll}
\hline
   \ \ \ \ \ \ \ \  \ \ \ \ \ \ \ \ \ \ \ \ Language Models \\
   \hline
    OpenAI   & \ \ \ \ \ ChatGPT\\
    Google       &  \ \ \ \ \ LaMDA, BARD, PaLM, mT5\\
     Deepmind     &  \ \ \ \ \ Gopher, Chinchilla, Sparrow\\
     Anthropic   & \ \ \ \ \ Claude\\
     Baidu    &  \ \ \ \ \ Ernie 3.0 Titan, Ernie Bot\\
     Huawei & \ \ \ \ \ PanGu-Alpha\\
     Meta & \ \ \ \ \ OPT-IML, BlenderBot-3\\
     AI21Labs & \ \ \ \ \ Jurrasic-1\\
     LG AI Research & \  \ \ \ \ Exaone \\
     NVIDIA & \ \ \ \ Megatron-Turing NLG \\
   \hline
\end{tabular}
\end{table}
ChatGPT is a language model powered by GPT-3. These models are particularly designed for conversational tasks and pre-trained on various topics, and can guide in different tasks such as providing information and answering questions. LaMDA is a type of Transformer-based model and can assist in free-flowing conversations. BARD is a chatbot model to answer natural language questions with the help of NLP and Machine Learning. PaLM is a language model based on a few-shot learning approach to handle different tasks. Model mT5 is a text-to-text transformer model trained on the mC4 corpus.

Gopher is DeepMind’s language model and is relatively more efficient than existing large language models in different tasks such as answering questions and logical reasoning.
Chinchilla also works the same as Gopher. It uses relatively less computing for fine-tuning tasks. Sparrow is a chatbot designed by DeepMind to respond to users’ questions accurately and lessen the risk of unsafe and incorrect answers.
Claude is an Al-based conversational model powered by advanced natural language processing and trained using a Constitutional Al technique. The Ernie 3.0 model was developed by Baidu and Peng Cheng Laboratory trained on huge unstructured data and acquired state-of-the-art results in over 60 Natural Language Processing tasks. Ernie Bot is an AI-powered Chinese language model relatively similar to OpenAI’s ChatGPT, able to  language understanding, language generation, and text-to-image generation. PanGu-Alpha was designed by Huawei as a Chinese-language model equivalent to OpenAI’s GPT-3. It is highly accurate to complete several language tasks such as question answering, text summarization, and dialogue generation. OPT-IML is released by Meta as a pre-trained language model and fine-tuned for strengthening the performance on natural language tasks such as text summarization, translation, and question answering. 


\section{Architecture for Knowledge Graph Construction and Completion from language Models}

\begin{figure}[h!]
\centering
\includegraphics[width=1\textwidth]{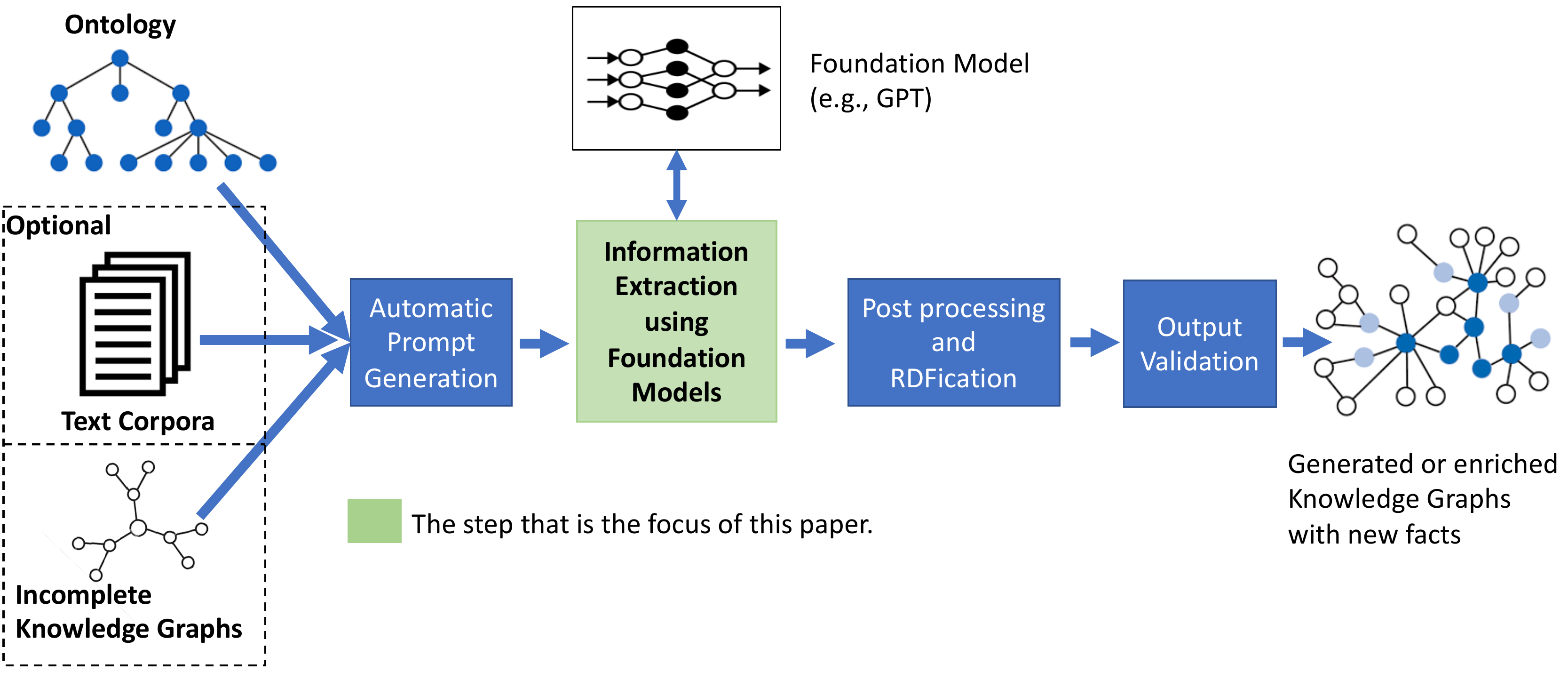}
\caption{A potential architecture for generating knowledge graphs with foundation models}
\label{fig:architecure}
\end{figure}

In this section, we present a potential architecture (see Figure~\ref{fig:architecure}) with a set of components for using foundational models for generating knowledge graphs from text corpora. It is important to note that even though we present this overall architecture, in the paper, our focus is only on one component, which is information extraction using the foundation model. The objective of presenting this architecture is to position our work in the overall big picture and motivate the usefulness of that work in a practical setting. We plan to work on the other components in future work to implement a pipeline for automatically generating knowledge graphs from text.

In the context of knowledge graph generation, we especially target on the following two possible scenarios: (a) knowledge graph construction from scratch and (b) knowledge graph completion where an incomplete knowledge graph is extended with missing facts. This pipeline aims to address both of these use cases.

\textbf{Inputs:} There are three possible inputs to the pipeline; an ontology, optionally text corpora, and an incomplete knowledge graph. The goal of the proposed pipeline is to generate a knowledge graph from text driven by an existing ontology, thus, one of the inputs to the process is the ontology. The ontology will guide which concepts and relations have to be used for information extraction for generating the triples. The facts can come from two sources: The knowledge that the foundation model acquired during its pre-training with a large volume of text (mainly open domain knowledge) or new text provided to it through the prompt (e.g., facts internal to a company from internal documents). In the latter case, the new text will come from a text corpora provided as input. In addition, an incomplete knowledge graph can also be provided as input. This will allow the pipeline to identify what are the missing facts and generate drive the information extraction to fill those missing information.

\textbf{Automatic Prompt Generation:}  Depending on the availability of computing resources and training data, there are different ways foundation models can be used for generating knowledge graphs from text by utilizing techniques such as fine-tuning, prompt tuning, or simple prompt designs. In the scope of this paper, we will focus only on the prompt design approaches. 

As the examples in Section 4 show, in order to perform the information extraction, we need to generate a prompt that provides instructions to the model on what needs to be extracted and also how to format the output. The goal of this component is to analyze the ontology and optionally the incomplete knowledge graph and generate prompts with concepts and relations to extract specific information (see Fig. \ref{fig:example_vaccione_manu}). This will require research related to prompt engineering~\cite{brown2020language, lester-etal-2021-power} as well as RDF profiling techniques~\cite{m2018rdf}.

\textbf{Information Extraction using Foundation Models} The goal of this component is to execute the generate prompts against the foundation model and generate the output. Some foundation models are publicly and freely available to download in repositories such as HuggingFace while some others are proprietary and made available through APIs based on a subscription. Depending on the use case, requirements, and the target foundation model, this component will have the necessary access mechanisms to run the inference of the model with a given input prompt.  

This component can be further improved using new paradigms that include teaching language models to use external tools such as Toolformer\cite{schick2023toolformer} or TALM~\cite{parisi2022talm}. This will allow the model to use other tools, for example, web search or other API calls necessary to complete the information request.

\textbf{Post-processing and RDFication:} As we can see in examples in Section 4 (see Fig. \ref{fig:example_vaccione_manu} or Fig. \ref{fig:example_4}), we are instructing the model to generate the facts in a simpler triplet format than a verbose RDF syntax. As the prompt and the response of the model are limited in the number of tokens\footnote{\url{https://help.openai.com/en/articles/4936856-what-are-tokens-and-how-to-count-them}}, this allows the response to be less verbose. The output of the model can then be post-processed and converted into RDF. As the pipeline has an ontology as input and the automatic prompt design component is aware of the concept and relations used to generate the prompt, this process can be done in a deterministic manner. 

\textbf{Output Validation:} As we will discuss in Section 5, one of the challenges in this process is the fact that the language models could generate inaccurate or outdated facts due to reasons such as hallucinations, incomplete and outdated training data, or even due biases in the training data. Thus, it is important that the facts generated by the model are validated before they are used in the knowledge graph. This is an open research question that requires further research. Certain validations can be done by performing reasoning with the generated triples to ensure there are no inconsistencies both at TBox and ABox levels. Another solution is to follow a Human-in-the-loop approach~\cite{monarch2021human,wu2022survey} or using crowd-sourcing if applicable~\cite{acosta2013crowdsourcing}.

\section{Qualitative analysis of information extraction capabilities of foundation models}\label{sec:evaluation}

This section provides a qualitative analysis of the information extraction capabilities of foundation models such as GPT using a set of examples that are based on the research questions. 

\textbf{R1: Can we use the knowledge acquired during pre-training of LLMs for Knowledge Graph completion with a defined ontology with a few examples?}

\begin{figure}[h!]
\caption{An example prompt to extract facts from the knowledge the model acquired during pre-training about manufacturers of COVID-19 vaccines.}
\centering
\includegraphics[width=1\textwidth]{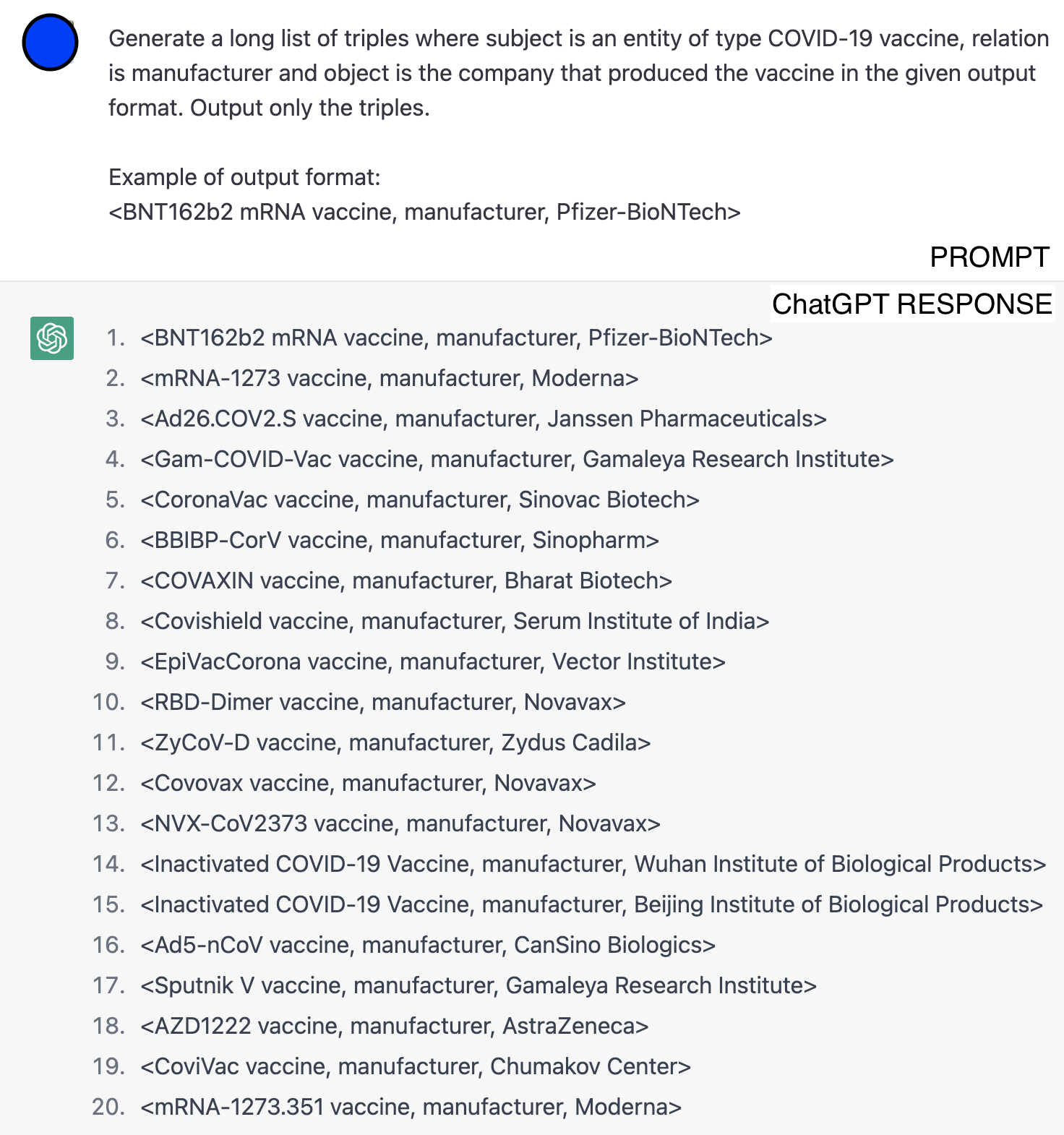}
\label{fig:example_vaccione_manu}
\end{figure}

\begin{figure}[h!]
\caption{Another example prompt to extract facts from the model about the COVID-19 transmission process.}
\centering
\includegraphics[width=1\textwidth]{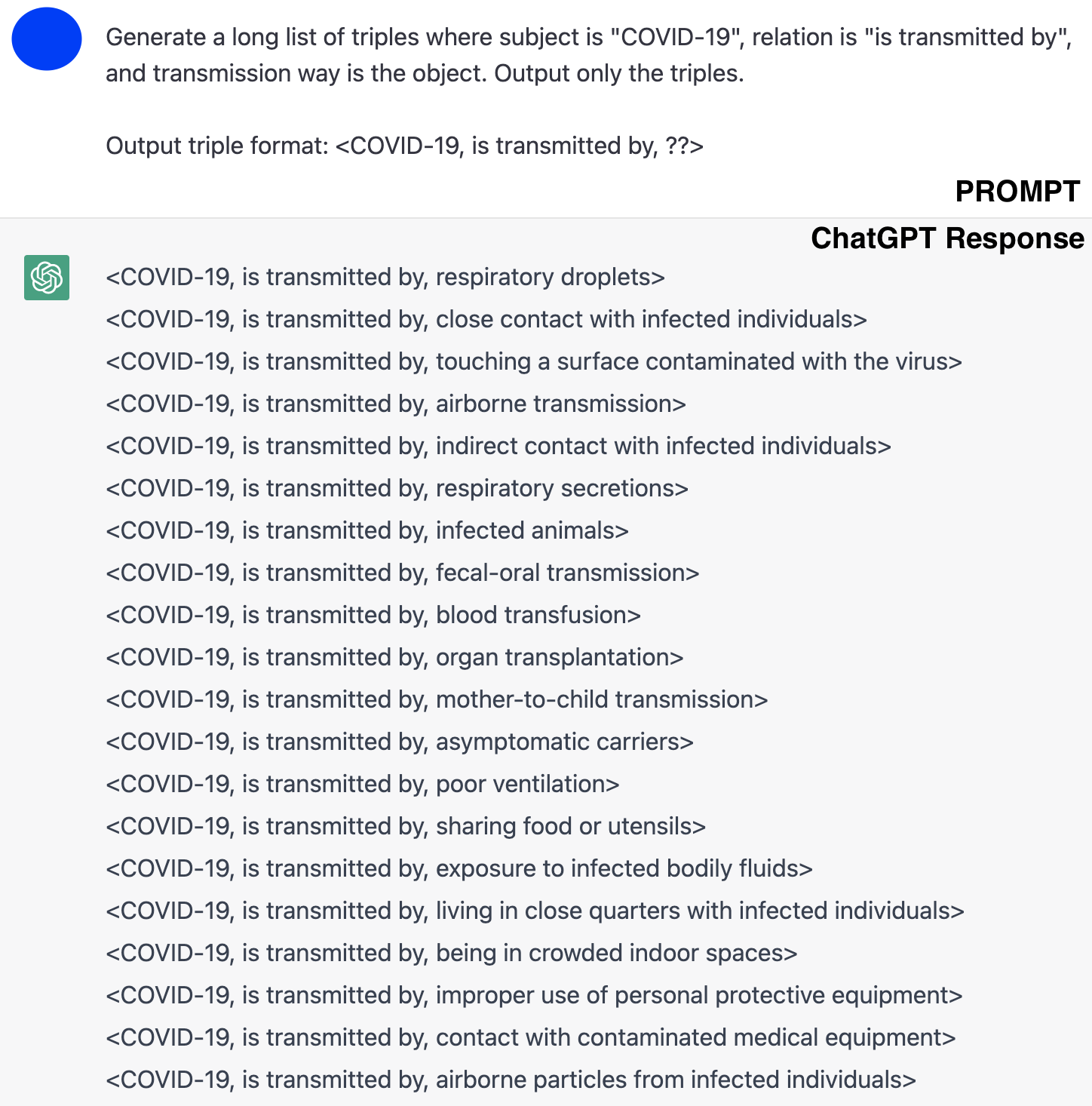}
\label{fig:example_4}
\end{figure}

Since most Covid-19 KGs have dealt with biomedical aspects of the disease and there has been less effort on the societal, economic, and climate change impacts of the disease\cite{Groppe2022AnalysisCOVID19}, We can ask Chatgpt if it is able to fill in the missing information in the knowledge graphs. Figure~\ref{fig:example_vaccione_manu} illustrates the use case that is studied in this question. It asks the model to generate a set of triples about COVID-19 vaccines and their manufacturers based on the knowledge that it has acquired during the pre-training. The output shows that the model understood the task and generated the triples following the output format based on the instructions in the prompt. Out of the 20 triples produced by the model, only one is factually inaccurate: The tenth triple is incorrect because ZF2001, the vaccine based on RDB-Dimer, is manufactured by Anhui Zhifei Longcom, in collaboration with the Institute of Microbiology at the Chinese Academy of Sciences and not by Novavax. The model might have made this mistake because the Wikipedia page states, `ZF2001 employs technology similar to other protein-based vaccines in Phase III trials from Novavax, Vector Institute, and Medicago'\footnote{https://en.wikipedia.org/wiki/ZF2001}. Figure~\ref{fig:example_4} illustrated another example with the entity ``COVID-19'' and the relation ``is transmitted by''. Also for this example, the model was able to find information that is not currently available in Wikidata.

Similarly, Table~\ref{tab:rq1} shows results for similar exercises on 10 other examples. For each example, we selected an entity and a relation from Wikidata and generated a prompt similar to the one in Fig.~\ref{fig:example_vaccione_manu}. Then each of the generated triples is manually checked to see if they are factually correct. Except for some facts in three distinct requests, all the other facts generated by the model were factually correct. But it must be mentioned that on the COVID-19 pandemic impact on tourism (Q9084098) entity, there have been several redundancies formulated by chatgpt, like 'Reduction in tourism spending' and 'Decrease in tourism spending' or 'Loss in revenue for airlines' and 'Decrease in airline revenue' which tend to be the same stuff but has been stated as two different entities by chatgpt. It is worth noting that in the third example, Prevention of SARS-CoV-2/COVID-19 (Q102056722), chatgpt provided some items that act as therapies but can be concluded also as preventative measures.

\begin{figure}[h!]
\caption{An example prompt to extract facts from an unstructured natural language sentence. Here we want the model to extract information only from the given text.}
\centering
\includegraphics[width=0.8\textwidth]{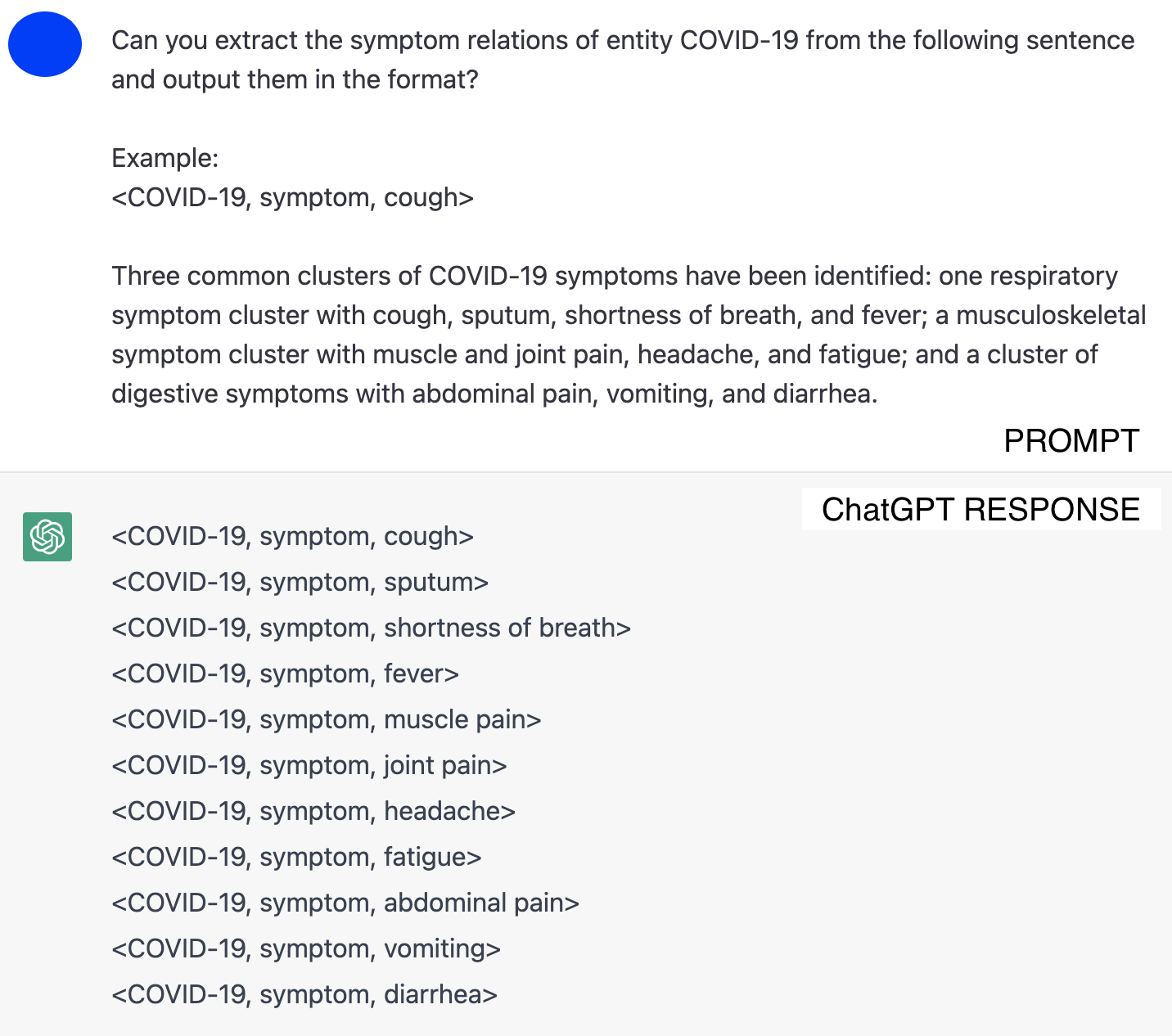}
\label{fig:example_symptom}
\end{figure}

\begin{table}[h!]
\centering
\caption{Summary of triples generated for 10 requests using prompts similar to Fig. \ref{fig:example_vaccione_manu} for completing missing information. Entities and relations are taken from Wikidata.}
\begin{tabular}{|p{10.5cm}|l|l|}
\hline
Requested Information & \begin{tabular}[c]{@{}l@{}}\# of triples\\ generated\end{tabular} & \begin{tabular}[c]{@{}l@{}}\# of correct\\ triples\end{tabular} \\ \hline
Subject Entity type: COVID-19 vaccine (Q87719492) \newline Relation: manufacturer (P176) & 20 & 19 \\ \hline
Subject Entity: vaccine hesitancy (Q56641686) \newline Relation: has contributing factor (P1479) & 15 & 15 \\ \hline
Subject Entity: Prevention of SARS-CoV-2/COVID-19 (Q102056722) \newline Relation: has part(s) (P527) & 26 & 26 \\ \hline
Object Entity: variant of SARS-CoV-2 (Q104450895) \newline Relation: instance of (P31) & 15 & 13 \\ \hline
Entity: COVID-19 Personal Protective Equipment During the Pandemic (Q100433360) \newline Relation: instance of (P31) & 4 & 4 \\ \hline
Entity: long COVID (Q100732653) \newline Relation: symptoms and signs (P780) & 13 & 13 \\ \hline
Entity: Covid-19 impact on pregnant women (Q117032167) \newline Relation: symptoms and signs (P780) & 6 & 6 \\ \hline
Entity: COVID-19 misinformation (Q85173778) \newline Relation: instance of (P31) & 6 & 6 \\ \hline
Entity:  COVID-19 pandemic impact on tourism (Q90840989) \newline Relation: instance of (P31) & 26 & 22 \\ \hline
Entity:  impact of the COVID-19 pandemic on the environment (Q90085156) \newline Relation: instance of (P31) & 20 & 20 \\ \hline
\end{tabular}
\label{tab:rq1}
\end{table}

\begin{table}[h!]
\centering
\caption{Summary of triples generated for 10 natural language triple extraction using prompts similar to Fig. 3.}
\begin{tabular}{|p{10cm}|l|l|}
\hline
Requested Information & \begin{tabular}[c]{@{}l@{}}\# of facts\\ in the sentences\end{tabular} & \begin{tabular}[c]{@{}l@{}}\# of facts\\ extracted  \end{tabular} \\ \hline
Entity: COVID-19 (Q84263196)\newline Relation: symptoms and signs (P780)  & 11 & 11 \\ \hline
Entity: COVID-19 (Q84263196)\newline
Relation: treatment (Q179661)
 & 8 &  5 \\ \hline
Entity: Long-term Effects of COVID-19 (Q113939303) \newline Relation: symptoms and signs (P780) & 15 & 13 \\ \hline
Entity: COVID-19 vaccine (Q87719492) \newline Relation: side effect (P1909) &  4 & 4\\ \hline
Entity: Covid-19 in children (Q97189089)\newline Relation: symptoms and signs (P780) & 10 &  10\\ \hline
Entity: Economic impact of the COVID-19 pandemic (Q96175652) \newline Relation: instance of (P31)
 & 21 & 21 \\ \hline
impact of the COVID-19 pandemic on religion (Q87898060)
\newline Relation: instance of (P31)
 & 6 & 6 \\ \hline
 COVID-19 related shortage (Q88429117)
\newline Relation: instance of (P31)
 & 8 & 7 \\ \hline
 COVID-19 disease in pregnancy (Q88058672)
\newline Relation: effect (Q926230)
 & 9 & 8 \\ \hline
long COVID (Q100732653)
\newline Relation: symptoms and signs (P780)
 & 12 & 12 \\ \hline
\end{tabular}
\label{tab:rq2}
\end{table}

\textbf{R2: Can we use LLMs to extract facts to generate knowledge graphs from unseen text that is provided during inference time?}

 In Figure \ref{fig:example_symptom}, an example prompt has been submitted to see if chatgpt is able to extract entities and relations  from raw text provided during inference. The prompt is about covid-19 symptoms and chatgpt extracted well all relations and entities.

Similar to the previous research question, Table~\ref{tab:rq2} shows an analysis of a set of examples following this second use case. 
Some entities and relations were taken from wikidata. Mostly there are lots of missing relations in wikidata for covid-19 related items. We provided some related text for chatgpt and asked that to make triples. As the table illustrates chatgpt acted well. In the second example, only 5 out of 8 triples were detected and chatgpt could not extract the treatments from the sentence  ’Other corticosteroids, such as prednisone, methylprednisolone or hydrocortisone, may be used if dexamethasone isn’t available.’
And in the third example, the long-term effects of  COVID-19, the following sentence was not detected by chatgpt ’Other symptoms were reported, which were not included in the publications, including brain fog and neuropathy'. It might be the case that chatgpt was not able to extract the triples as it is stated in the phrase "which were not included in the publications". On example eight, chatgpt was only able to extract 7 out of 8 facts. There has been a sentence in the text 'In some cases, governmental decision making created shortages, such as when CDC prohibited the use of any diagnostic test other than the one it created.'. Chatgpt extracted the following wrong triple: <diagnostic test other than the one it created, instance of, COVID-19 related shortage>. the term 'when' is vital to take into consideration, although the last sentence has the phrase 'created shortages, such as'. In the ninth example, COVID-19 disease in pregnancy, chatgpt has detected the wrong triple on the sentence, 'A review in 2022 suggests that pregnant women are at increased risk of severe COVID-19 disease, with an increased rate of being hospitalized to the intensive care unit and requiring ventilation, but was not associated with a statistically significant increase in mortality.'. The wrong triple is 'COVID-19 disease in pregnancy-effect-mortality'. Chatgpt neglected the phrase 'was not associated with'. One other stuff to take into account is the fact that prompt design is really crucial in extracting triples by chatgpt. It is essential to provide an example triple from the same text that is given to chatgpt.

\textbf{R3: Given an ontology, can we automatically generate prompts for extracting the relevant triples for the purpose of Knowledge Graph construction?}\\

\begin{figure}[h!]
\caption{An example prompt to generate triples given an ontology as input in the prompt.}
\centering
\includegraphics[width=0.8\textwidth]{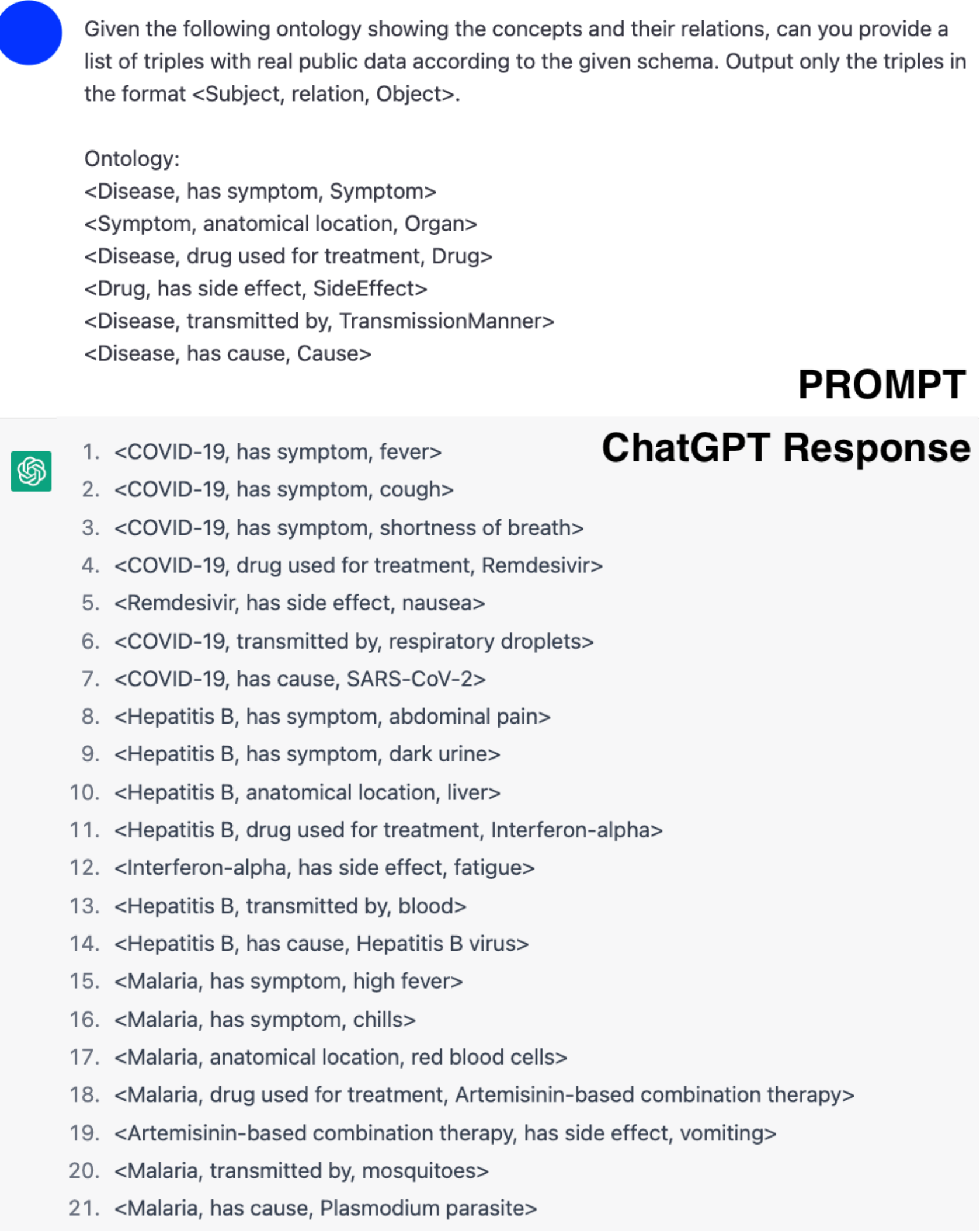}
\label{fig:example3}
\end{figure}

Figure~\ref{fig:example3} shows an example of this use case. We have provided a prompt with a toy ontology about diseases which contains 7 concepts such as disease, symptom, organ, drug, and 6 relations with their domain and range concepts. As the example's output shows, the model seemed to understand the task and provided an output with a set of triples that followed the given ontology. In some triples, it deviated from the given domain constraints of the relation, for example, in triples 10 and 17, it provided ``anatomical location'' for the disease instead of the symptom. Nevertheless, the extracted facts are correct in those cases. Overall, the generated triples seem to follow the schema and are factually correct. 

Because there are token limits for both the prompt input as well as the output, in order to follow a similar setup for a larger real-world ontology, we would have to perform an iterative process. As the initial results for the toy example have shown promising results, in future work, we will follow up with a larger ontology to extract a list of triples to construct a knowledge graph given an ontology.

\textbf{R4: Given a knowledge graph, can we identify the missing information and create prompts to perform Knowledge Graph completion using foundation models}\\
\begin{figure}[h!]
\caption{An example prompt to generate relations for an existing entity.}
\centering
\includegraphics[width=0.8\textwidth]{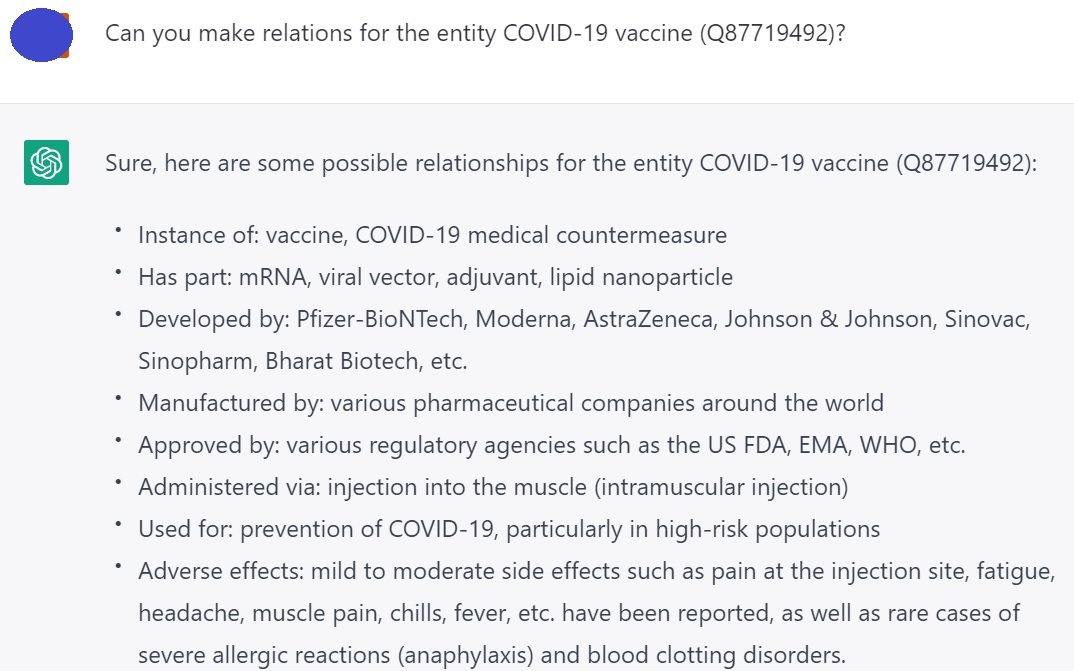}
\label{fig:RQ4-example}
\end{figure}
There might be missing entities or links in a knowledge graph. This missing info can be predicted by the entity and link prediction methods, then we can design prompts and feed that to LLMs. So  we can pick an entity that misses links and ask the foundation model to provide us with related relations for the entity as depicted in Figure \ref{fig:RQ4-example}.

\textbf{R5: Given a Knowledge Graph  with some false facts, can we use LLMs to check the given Knowledge Graph and determine which facts are not true for the purpose of fact-checking?\\}
\begin{figure}[h!]
\caption{An example prompt for fact-checking.}
\centering
\includegraphics[width=0.8\textwidth]{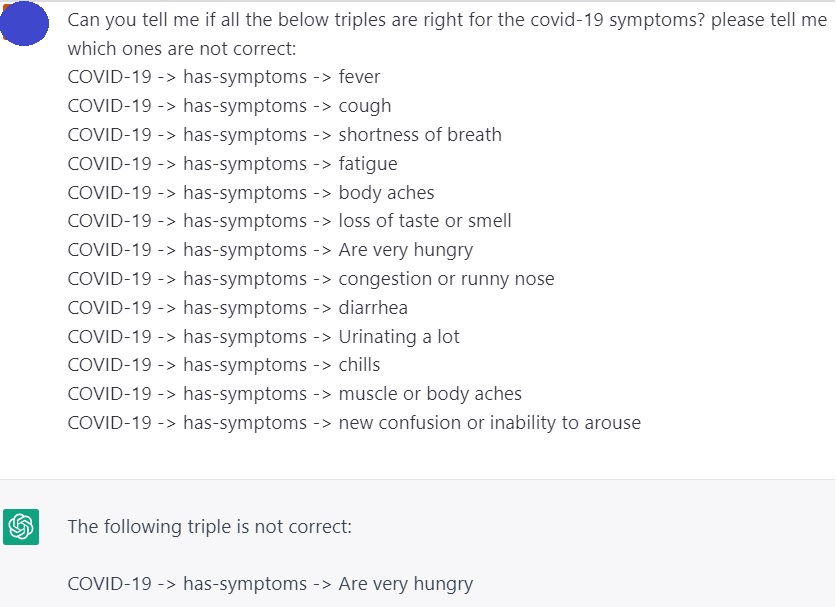}
\label{fig:RQ5-example}
\end{figure}
That task is a bit tricky with chatgpt as it sometimes hallucinates and provides wrong data. But overall it is possible to compare what chatgpt provides as relations and objects and compare with what is already in the knowledge graph. As Figure \ref{fig:RQ5-example} shows chatgpt only found 'Are very hungry' and neglected the wrong fact 'urinating a lot'.
However, if the training data is biased, chatgpt might provide some wrong output. Or if doesn't have access to the data, like if the knowledge graph contains some data or statistics that belong to a certain company then it would provide imprecise results. So although LLMs could be considered a valuable asset in fact checking, they must be supplemented with some other methods for better accuracy and precision. 

\section{Discussion}

In this section, we will discuss the advantages of having an automatic Knowledge Graph construction/completion pipeline and also reflect on the current challenges. These challenges open the door to future research to mitigate them and make such automatic knowledge graph construction pipeline more robust. 

\subsection{Advantages}


\subsubsection{Cost-effective Knowledge Graph Creation and Completion}

As shown in Section~\ref{sec:evaluation}, foundation models inherently contain a large amount of information that is not yet available as structured data. If we need to fill that information with the help of domain experts or crowd-sourcing, it requires a vast amount of human effort.  There are high efforts needed causing high costs to generate or complete knowledge graphs based on large-scale text collections. It seems to be less cost-intensive and faster to use pre-trained language models for knowledge graph creation and completion.

\subsubsection{Scalability} 
If we can employ an automatic pipeline for knowledge base creation and construction with minimal human effort, we can scale the creation and completion of the knowledge graphs just with computational resources. If humans have to manually construct and curate the knowledge graphs, it is not scalable, especially for custom knowledge graphs, for example, for a given organization. 

\subsubsection{Continuous updates for evolving knowledge graphs}
If we can automatically convert unstructured natural language text to knowledge graphs, we will be able to convert the most up-to-date information coming from sources such as news articles or social media posts into facts in knowledge graphs.

\subsubsection{Size of Language Models versus Raw Text for Knowledge Graph Creation and Completion}
The size of language models is less than and relatively compact compared to the original raw texts used for pre-training of the language models. Hence it is easier to set up a system for knowledge graph construction and completion using pre-trained language models instead of natural language processing of large-scale text collections. 

If the large-scale text collections are not available for download in one (or a set of) compressed files, then a crawler is needed to retrieve the text collection increasing the technical complexity and the processing time. Compared to using crawlers, if the language model cannot be installed locally, then it is relatively easy to communicate with the corresponding chatbot by generating the prompts and processing the answers for knowledge graph creation and completion by using typically a well-defined web API of the chatbot.

\subsection{Challenges}

\subsubsection{Hallucinations}

Foundation models and large language models are known to create hallucinations when they generate natural language where they invent non-existing facts (also referred to as being unfaithful to the source content) or nonsensical text in a fluent and confident manner~\cite{ji2022survey}. This phenomenon has been observed in many language generation tasks such as conversational dialogues, text summarization, generative question answering, data-to-text generation and machine translation. For a human or a downstream system, it can be hard to identify which outputs of the model are non-factual. It can have undesired outcomes, and failures in the downstream tasks if the hallucinations can not be detected and filtered out. This could result in bad user experiences in real-world applications. 

\subsubsection{Bias and Fairness}

Though the large language models have recently shown impressive results on many academic benchmarks for various NLP tasks, it is still less understood what different types of biases exist in these models~\cite{kwon2022-empirical}. As the training data could reflect societal biases that discriminate against certain groups of individuals in an unfair manner, the outputs of these models could be susceptible to such biases. NLP research communities are working towards defining methods to uncover such biases in large language models and mitigating them ensuring both individual and group fairness of the results. If not addressed properly, knowledge graphs constructed using these language models could inherit some of the biases from these models.

\subsubsection{High Computational Resources}

Foundational models are inherently large in their size. For example, OpenAI's ChatGPT has 175B parameters, Meta's Galactica has 120B parameters, and Googgle's Bard 137B parameters, BigScience's BLOOM 176B parameters, and so on. Thus, these models require high computation resources to run including GPUs, and large memories.

\subsubsection{Automatic Prompt Design}
One of the requirements of these instruction-based foundation models is to come up with prompts for all the information that needs to be extracted. For example, if we have an ontology (TBox) and want to populate a knowledge graph with facts (ABox), it will require generating correct prompts and optionally examples or demonstrators. Further research and exploration are needed to understand if that can be easily done through templates.

\section{Conclusions and Future Work}

In the era of rapid advancements in large language models  or foundation models and their applications, it is important to explore how these large language models can be used to improve knowledge graphs and inversely how Knowledge Graphs can be used to improve large language models. In this work, we have evaluated a large foundation model, i.e., GPT-3.5 using ChatGPT, for the purpose of understanding its capabilities for tasks related to knowledge graph construction and completion. 
We made a qualitative analysis of ChatGPT on knowledge graph construction and completion based on research questions discussed earlier in the paper. As the results show ChatGPT can be considered a valuable source in knowledge graph construction and completion but we should keep in mind that there are some challenges including bias, hallucinations, and high computational costs. Another stuff that needs to be considered is the fact that prompt design is an extremely important matter in this area. As improper prompt design might lead to inaccurate results.

In future work, we plan to implement an automatic pipeline that uses foundation models to perform information extraction and generate knowledge graphs from text. There are several open research challenges that need to be addressed to accomplish this including automatic prompt generation using ontologies and validation of the generated output. We believe complementary use of foundation models and knowledge graphs opens up several new research directions and we plan to further explore each of these areas in the context of knowledge graph generation from text.

\bibliography{ceur}
\end{document}